\documentclass[sigconf]{acmart}

\usepackage{booktabs} % For formal tables
\usepackage{multirow}  % [MY]
\usepackage{amsmath}  % [MY]
\usepackage{balance}  % [MY]

\setcopyright{rightsretained}
\copyrightyear{2018} 
\acmYear{2018} 
\setcopyright{acmcopyright}
\acmConference[KDD '18]{The 24th ACM SIGKDD International Conference on Knowledge Discovery \& Data Mining}{August 19--23, 2018}{London, United Kingdom}
\acmBooktitle{KDD '18: The 24th ACM SIGKDD International Conference on Knowledge Discovery \& Data Mining, August 19--23, 2018, London, United Kingdom}
\acmPrice{15.00}
\acmDOI{10.1145/3219819.3220036}
\acmISBN{978-1-4503-5552-0/18/08}

\begin{document}
\title{R-VQA: Learning Visual Relation Facts with Semantic Attention for Visual Question Answering}

\author{Pan Lu}
\authornote{This work was mainly performed when Pan Lu was visiting Microsoft Research.}
\affiliation{%
	\department{Dept. of Computer Science}
  	\institution{Tsinghua University}
}
\email{lupantech@gmail.com}

\author{Lei Ji$^{1,2}$}
\affiliation{%
	\institution{Microsoft Research Asia$^{1}$, ~Institute of Computing Technology, CAS$^{2}$}
}
\email{leiji@microsoft.com}

\author{Wei Zhang}
\authornote{Wei Zhang is the corresponding author.} 
\affiliation{%
	\department{School of CSSE}
	\institution{East China Normal University}
}
\email{zhangwei.thu2011@gmail.com}

\author{Nan Duan}
\affiliation{%
	\department{Microsoft Research Asia}
	\institution{Microsoft Corporation}
}
\email{nanduan@microsoft.com}

\author{Ming Zhou}
\affiliation{%
	\department{Microsoft Research Asia}
	\institution{Microsoft Corporation}
}
\email{mingzhou@microsoft.com}

\author{Jianyong Wang}
\affiliation{%
	\department{Dept. of Computer Science}
	\institution{Tsinghua University}
}
\email{jianyong@mail.tsinghua.edu.cn}

\renewcommand{\shortauthors}{P. Lu et al.}

\begin{abstract}

Recently, Visual Question Answering (VQA) has emerged as one of the most significant tasks in multimodal learning as it requires understanding both visual and textual modalities. Existing methods mainly rely on extracting image and question features to learn their joint feature embedding via multimodal fusion or attention mechanism. Some recent studies utilize external VQA-independent models to detect candidate entities or attributes in images, which serve as semantic knowledge complementary to the VQA task. However, these candidate entities or attributes might be unrelated to the VQA task and have limited semantic capacities. To better utilize semantic knowledge in images, we propose a novel framework to learn visual relation facts for VQA. Specifically, we build up a Relation-VQA (R-VQA) dataset based on the Visual Genome dataset via a semantic similarity module, in which each data consists of an image, a corresponding question, a correct answer and a supporting relation fact. A well-defined relation detector is then adopted to predict visual question-related relation facts. We further propose a multi-step attention model composed of visual attention and semantic attention sequentially to extract related visual knowledge and semantic knowledge. We conduct comprehensive experiments on the two benchmark datasets, demonstrating that our model achieves state-of-the-art performance and verifying the benefit of considering visual relation facts.

\end{abstract}

% https://dl.acm.org/ccs.cfm
% https://dl.acm.org/ccs/ccs_flat.cfm
\begin{CCSXML}
	<ccs2012>
	<concept>
	<concept_id>10002951.10003317.10003347.10003348</concept_id>
	<concept_desc>Information systems~Question answering</concept_desc>
	<concept_significance>500</concept_significance>
	</concept>
	<concept>
	<concept_id>10010147.10010178.10010187</concept_id>
	<concept_desc>Computing methodologies~Knowledge representation and reasoning</concept_desc>
	<concept_significance>500</concept_significance>
	</concept>
	</ccs2012>
\end{CCSXML}

\ccsdesc[500]{Computing methodologies~Knowledge representation and reasoning}
\ccsdesc[300]{Information systems~Question answering}

\keywords{visual question answering; relation fact mining; semantic knowledge; attention network; question answering}

\maketitle

\section{Introduction}

\begin{figure}[t]
	\centering
	\includegraphics[width=0.95\linewidth]{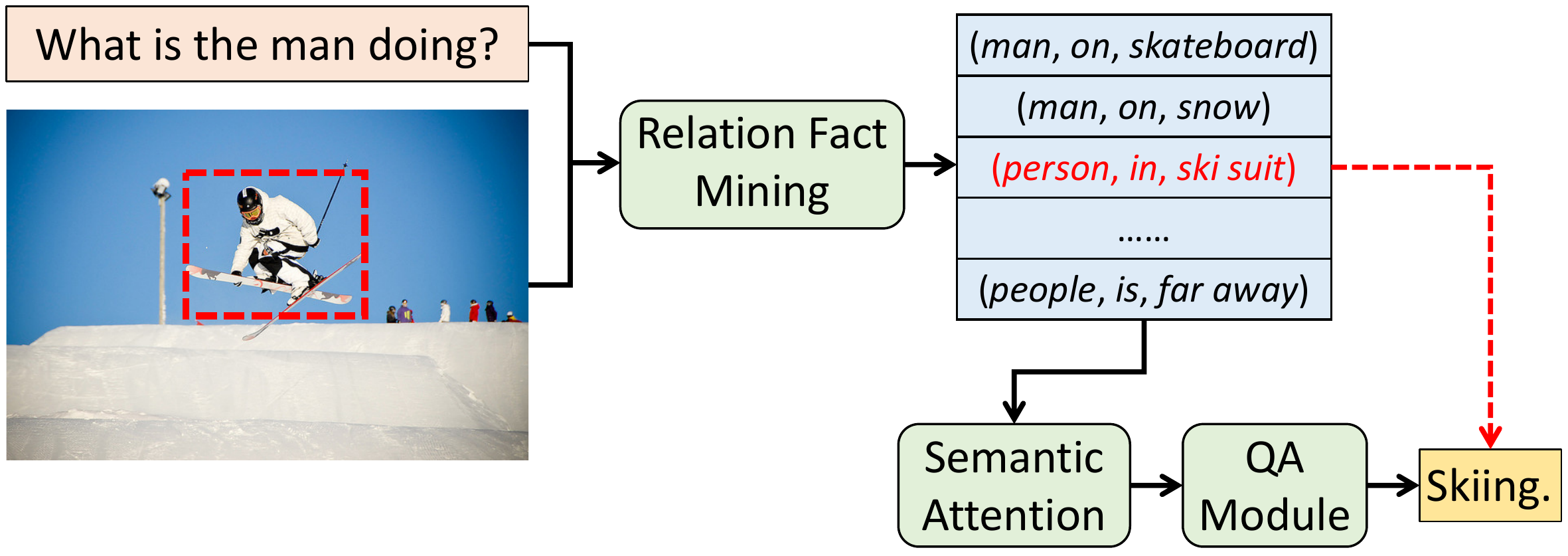}
	\caption{Our proposed model, which learns to mine relation facts with semantic attention for visual question answering.}
	\label{example}
\end{figure}
\vspace{-0.5mm}

With the great development of natural language processing, computer vision, knowledge embedding and reasoning, and multimodal representation learning, Visual Question Answering has become a popular research topic in recent years. The VQA task is required to provide the correct answer to a question with a corresponding image, which has been regarded as an important Turing test to evaluate the intelligence of a machine. The VQA problem can be easily expanded to other tasks and play a significant role in various applications, including human-machine interaction and medical assistance. However, it is difficult to address the problem, as the AI system needs to understand both language and vision content, to extract and encode necessary common sense and semantic knowledge, and then to make reasoning to obtain the final answer. Thanks to multimodal embedding methods and attention mechanisms, researchers have made remarkable progress in VQA development.

The predominant methods first extract language feature embedding by an RNN model and image feature embedding by a pre-trained model, then learning their joint embedding by multimodal fusion like element-wise addition or multiplication, and finally feeding it to a sequential network to generate free-form answers or to a multi-class classifier to predict most related answers. 
Inspired by image captioning, some VQA approaches \cite{wu2016ask,li2017incorporating,wu2017image} introduce semantic concepts such as entities and attributes from off-the-shelf CV methods, which provide various semantic information for the models.
Compared with entities and attributes, relation facts have larger semantic capacities as they consist of three elements: subject entity, relation, and object entity, leading to a large number of combinations. For example, in Figure \ref{example}, given the question "what is the man doing" and the image, relation facts like \textit{(man, standing on, skateboard)},  \textit{(man, on, ice)}, \textit{(person, in, ski suit)} enable providing important semantic information for question answering.

The main challenge for VQA lies in the semantic gap from language to image. To deal with the semantic gap, existing attempts come in
two forms. To be specific, some methods extract high-level semantic information \cite{wu2016ask,li2017incorporating,wu2017image}, such as entities, attributes, or even retrieval results in knowledge base \cite{li2017incorporating}, such as DBpedia \cite{auer2007dbpedia} and Freebase \cite{bollacker2008freebase}. Other methods introduce visual attention \cite{xu2016ask,yang2016stacked,kim2016hadamard} to select related image regions corresponding to salient visual information. 
Unfortunately, these progressions of introducing semantic knowledge are still limited in two aspects. On one hand, they use entities or attributes as high-level semantic concepts, which are individual and only cover restricted knowledge information. On the other hand, as they extract the concept based on off-the-shelf CV methods in other tasks or datasets, the candidate concepts might be irrelevant to the VQA task.

To make full use of  semantic knowledge in images, we propose a novel semantic attention model for VQA. We build a large-scale Relation-VQA (R-VQA) dataset including over 335k data samples based on the Visual Genome dataset. Each data instance is composed of an image, a relevant question, and a relation fact semantically similar to the image-question pair. We then adopt a relation detector to predict the most related visual relation facts given an image and a question. We further propose a novel multi-step attention model to incorporate visual attention and semantic attention into a sequential attention framework. 
%As shown in Figure \ref{vqamodel}, our model is composed of three major components. 
Our model is composed of three major components (see Figure \ref{vqamodel}). 
The visual attention module (Subsection \ref{subsection:51}) is designed to extract image feature representation. The output of the visual attention module is then fed into semantic attention (Subsection \ref{subsection:52}), which learns to select important relation facts generated by the relation detector (Section \ref{section:4}). Finally, joint knowledge learning (Subsection \ref{subsection:53}) is applied to simultaneously learn visual knowledge and semantic knowledge based on visual and semantic feature embeddings.

The main contributions of our work are four-fold. 
\begin{itemize}
	\item We propose a novel VQA framework which enables  learning visual relation facts as semantic knowledge to help answer the questions.   
	\item We develop a multi-step semantic attention network (MSAN) which combines visual attention and semantic attention sequentially to simultaneously learn visual and semantic knowledge representations.
	\item To achieve that, we build up a large-scale VQA dataset accompanied by relation facts and design a fine-grained relation detector model.
	\item We evaluate our model on two benchmark datasets and achieve state-of-the-art performance. We also conduct substantial experiments to illustrate the ability of our model.
\end{itemize}

%%%%%%%%%%%%%%%%%%%%%%%%%%%%%%%%%%%%%%%%%%%%
\section{Related Work}

\subsection{Visual Question Answering}
As the intersection of natural language processing, knowledge representation and reasoning, and computer vision, 
the task of Visual Question Answering has attracted increasing interest recently in multiple research fields. 
A series of large-scale datasets have been constructed, including VQA \cite{antol2015vqa},
COCO-QA \cite{ren2015exploring}, and Visual Genome \cite{krishna2017visual} datasets. 
A commonly used framework is to first encode each question as a semantic vector using a long short-term memory network (LSTM) 
and to extract image features via a pre-trained convolution neural network (CNN),
then to fuse these two feature embeddings to predict the answer.
In contrast to work in  \cite{noh2016image,ilievski2016focused,xiong2016dynamic} which use simple feature fusion like 
element-wise operation or concatenation, effective bilinear pooling methods are well studied in \cite{fukui2016multimodal,kim2016hadamard,benyounescadene2017mutan}.

\subsection{Attention Methods}
Attention networks have recently shown remarkable success in many applications of knowledge mining and natural language processing, such as neural machine translation  \cite{bahdanau2014neural}, recommendation systems \cite{wang2017dynamic}, advertising  \cite{zhai2016deepintent}, document classification \cite{yang2016hierarchical}, sentiment analysis \cite{long2017cognition}, question answering \cite{li2017context}, and others.
Bahdanau et al. \cite{bahdanau2014neural} introduced an attention mechanism to automatically select parts of words in a source sentence
relevant to predicting a target word, which improves the performance of basic encoder-decoder architecture. Long et al. \cite{long2017cognition}
propose a cognition based attention (CBA) layer for neural sentiment analysis to help capture the attention of words in source sentences.
Different from above works focusing on word-level attention, sentence-level \cite{yang2016hierarchical}
and document-level attention \cite{wang2017dynamic} pay more holistic attention to the whole textual content.
An attention mechanism has also been successfully applied to computer vision tasks like image captioning \cite{you2016image}, image retrieval \cite{liu2015content}, image classification \cite{xiao2015application}, 
image popularity prediction \cite{zhang2018user}, et al.

% For example, in image caption task, the algorithm is designed to generate language description about the salient visual information in given images. Xu et al. \cite{xu2015show} integrated attention mechanism in the RNN model in order to generate word sequence through dynamically focusing on relevant image regions based on generated words.

Inspired by the great success achieved by attention mechanisms on natural language processing and computer vision, 
lots of VQA approaches perform attention mechanism to improve model capacity.
Current attention methods \cite{xu2016ask,yang2016stacked} for VQA mainly perform visual attention
to learn image regions relevant to the question. Some recent works \cite{kim2016multimodal,fukui2016multimodal,kim2016hadamard,gao2018question}
integrate effective multimodal feature embedding with visual attention to further improve VQA performance. 
More recently, 	Lu et al. \cite{lu2018co-attending} design a novel dual attention network which introduces two types of visual features and enables learning question-releted free-form and detection-based image regions.
Different from these studies, we propose a novel sequential attention mechanism
to seamlessly combine visual and semantic clues for VQA.

\subsection{Semantic Facts}
Relation facts, standing for relationships between two entities, play an important role in representation and reasoning in knowledge graph. The encoding and applications of relation facts have been widely studied in multiple tasks of knowledge representation  \cite{socher2013reasoning,bordes2013translating}. Visual relationship detection is an emerging task aiming to generate relation facts\cite{lu2016visual, li2017vip}, e.g. \textit{(man, riding, bicycle)} and \textit{(man, pushing, bicycle)}, which capture various interactions between pairs of entities in images.

Existing relevant VQA methods involve using knowledge information to either obtain retrieval results of entities and
attributes \cite{wu2016ask,li2017incorporating,wu2017image}, or detect high-level concepts in the image according to the question query \cite{wu2016value}.
However, it is not effective enough to exploit the complicated semantic relations between the question and image
by simply treating semantic knowledge in images as entities and attributes.
To the best of our knowledge, it is still rare to incorporate relation facts in VQA to provide rich semantic knowledge.
In order to utilize relation facts in the VQA task, we propose effectively
learning relation facts and selecting related facts via semantic attention.

%%%%%%%%%%%%%%%%%%%%%%%%%%%%%%%%%%%%%%%%%%%%
\section{Preliminary}

In this section, we first formulate the VQA problem addressed in this paper, and then clarify the predominant framework for the problem.

\subsection{Problem Formulation}
Given a question $Q$ and a related image $I$, the VQA algorithm is designed to predict the most possible answer $\hat{a}$ based on both the language and image content. The predominant approaches formalize VQA as a multi-class classification problem in the space of candidate answer phrases from most frequent answers in training data. This can be formulated as
\begin{align}
	\hat{a} = \mathop{\arg\max}\limits_{a \in \Omega}  p(a|Q,I; \varTheta) ,
\end{align}
where $\varTheta$ denotes the parameters of the model and $\Omega$ is the set of candidate answers.

\subsection{Common Framework}
The common frameworks for VQA are composed of three major parts: the image embedding model, the question embedding model, and the joint feature learning model. CNN models like \cite{simonyan2014very,he2016deep} are used in the image model to extract image feature representation. For example, a typical deep residual network ResNet-152 \cite{he2016deep} can extract the image feature map $v$  from the last convolution layer before the pooling layer, which is given by:
\begin{align}
	v= \mathrm{CNN}(I) .
\end{align}
Before being fed into the CNN model, the input image is resized to be $448\times448$ pixels from the raw image.
The convolution feature map extracted from the CNN model has a size of $2048\times14\times14$, where $14\times14$ is its spatial size corresponding to different image regions, and $2048$ represents the number of feature embedding dimension of each region.

For the question model, recurrent neural networks like Long Short-Term Memory (LSTM) \cite{hochreiter1997long} and Gated Recurrent Unit (GRU) \cite{cho2014learning} are utilized to obtain the question semantic representation, which is given by:
\begin{align}
	q = \mathrm{RNN}(Q) .
\end{align}
To be specific, given a question with $T$ words, the embedding of each question word is sequentially fed into the RNN model. The final hidden state $h_T $ of the RNN model is taken as the question embedding.

The question and image representations are then jointly embedded into the same space through multimodal pooling, including element-wise product or sum, as well as the concatenation of these representations
\begin{align}
	h = \Phi(v, q) ,
\end{align}
where $\Phi$ is the multimodal pooling module. The joint representation $h$ is then fed to a classifier which predicts the final answer.

A large quantity of recent works incorporate visual attention mechanisms for more effective visual feature embedding. In general, a semantic similarity layer is introduced to calculate the relevance between the question and image regions defined as:
\begin{align}
	m_i = \mathrm{sigmoid}( \psi(q, v_i)) ,
\end{align}
where $\psi$ is the module of semantic similarity, $\mathrm{sigmoid}$ is a sigmoid-type of function, such as softmax, to map the semantic results to the value interval [0,1], and $m_i$ is the semantic weight of one image region. Finally, the visual representation of the image is updated by the weighted sum over all image regions as:
\begin{align}
	\tilde{v} =\sum_{i=1}^{14\times14} m_i v_i ,
\end{align}
which is able to highlight the representations of image regions most related to the input question.

%%%%%%%%%%%%%%%%%%%%%%%%%%%%%%%%%%%%%%%%%%%%
\section{Relation Fact Detector} \label{section:4} 
In this section, we describe the process of collecting our Relation-VQA (R-VQA) dataset,
as well as the data analysis in Subsection \ref{subsection:41}.
We then design a relation fact detector in Subsection \ref{subsection:42} based on R-VQA to predict
visual relation facts related to given questions and images,
which is further incorporated into our VQA model in Section \ref{section:vqa}.

\subsection{Data Collection for Relation-VQA} \label{subsection:41} 
\textbf{Survey of Existing Datasets~} Existing VQA datasets like VQA \cite{antol2015vqa} and COCO-QA \cite{ren2015exploring}
are made up of images, questions and labeled answers, not involving supporting semantic relation facts.
Although the Visual Genome Dataset \cite{krishna2017visual} provides semantic knowledge information such as objects,
attributes, and visual relationships about parts of images, which is not aligned with their corresponding question-answer pairs.
Therefore, we expand Visual Genome based on semantic similarity and build up the Relation-VQA dataset,
which is composed of questions, images, answers, and aligned semantic knowledge. 
The dataset will be released at 
%\textcolor[RGB]{122,122,122}{
%\url{https://github.com/lupantech/rvqa}}
%.  
{\color{magenta}{
	\url{https://github.com/lupantech/rvqa}}
}\!.

% tab1:rvqa-triplet
\begin{table}[h]
	\centering 
	%\small 
	\footnotesize
	\begin{tabular}{*3{c}} 
		\toprule	
		Semantic Knowledge				& Fact Templates	 		& Examples   \\
		\midrule
		Entity concept	& (there, is, object)	 			& (\textit{there, is, train station} )  \\
		Entity attribute	& (subject, is, attribute)	 		& (\textit{plate, is, white})   \\
		Entities' relation  & (subject, relation, object)	 	& (\textit{computer, under, desk})  \\
		\bottomrule		
	\end{tabular}
	\caption{Types of relation facts.}
	\label{tab:rvqa-triplet}
\end{table}
%\vspace{-0.5mm}

\textbf{Data Collection~}  We first define relation facts used in our paper as shown in Table \ref{tab:rvqa-triplet}.
The  relation facts are categorized as one of three types:
\textit{entity concept}, \textit{entity attribute}, and \textit{entities relation},
based on the semantic data of concepts, attributes, and relationships in Visual Genome, respectively.
For simplicity, the \textit{attribute} in an entity attribute can be an adjective, noun, or preposition phrase.
For Visual Genome, most images are provided with related question-answer pairs, and parts of images are annotated with semantic knowledge.
Thus, we keep images containing both question-answer pairs and semantic knowledge,
and treat these semantic knowledge as candidate facts with the form of the above templates.

Response Ranking, a semantic similarity ranking method proposed in \cite{yan2016docchat}, 
is then adopted to compute the relevance between each QA pair and its candidate facts.
It should be noted that as the ranking algorithm is not our work's main focus and various ranking algorithms are compatible in our framework, we simply adopt one of the state-of the-art ranking methods, such as Response Ranking. We leave the choice or design of a better ranking algorithm in future work.
The relevance matching score obtained from the Response Ranking module ranges from 0 to 1, and value 0 means the candidate fact is
completely unrelated to the given QA data, while value 1 means perfect correlation.
In the end, after removing candidate facts below a certain threshold matching score,
the fact with the largest score is chosen as the ground truth.
We randomly partition the generated data into a training set (60\%), a development set (20\%) and a testing set (20\%).
Table \ref{tab:rvqa-data} shows the data sizes of R-VQA with different matching score thresholds.

% tab2:rvqa-data
\begin{table}[ht]
	\centering 
	%\small 
	\footnotesize
	\begin{tabular}{*7{c}} 
		\toprule	
		Score THR	& \# Train 	& \# Dev 	& \# Test  	& \# Total 	& \# Unique Img. & Match  \\
		\midrule
		0.20		& 286,972	& 95657 	& 95658 	& 478,287 	& 78,863    & 75.6\% 	\\		
		0.25		& 207,589	& 69,196	& 69,198 	& 345,983  	& 72,993 	& 86.1\% 	\\	
		0.30		& 119,333	& 39,777 	& 39,779 	& 198,889   & 60,473  	& 90.8\% 	\\	
		0.35		& 28,345	& 9,448 	& 9,449 	& 47,242    & 25,884	& 91.7\% 		\\	
		0.40		& 24,668	& 8,222		& 8,224		& 41,114	& 23,480 	& 93.1\% 	\\	
		0.45		& 756		& 252		& 253		& 1,261  	& 1,096  	& 95.0\% 	\\	
		\bottomrule		
	\end{tabular}
	\caption{Basic statistics of the R-VQA dataset.}
	\label{tab:rvqa-data}
\end{table}
% \vspace{-1mm}

\begin{figure}[t]
	\centering
	\includegraphics[width=0.95\linewidth]{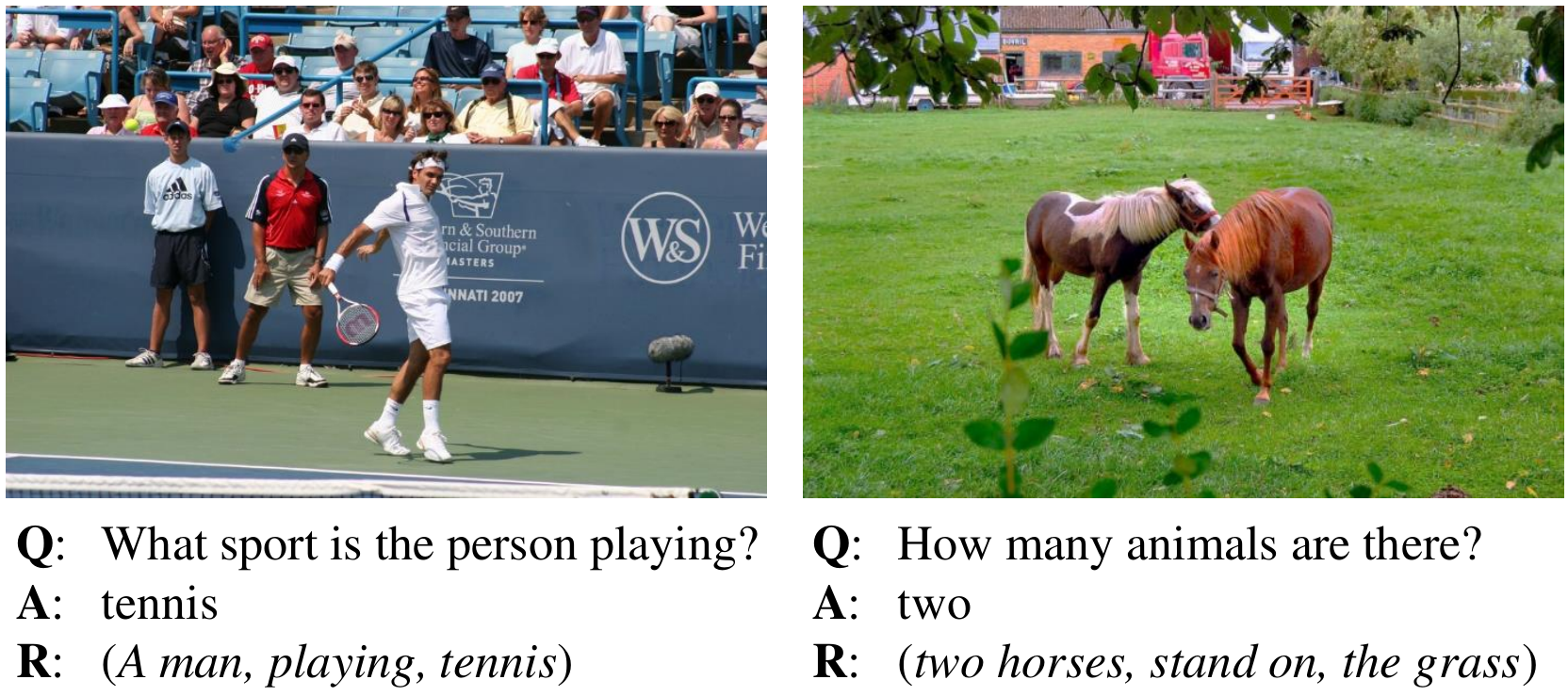}
	\caption{Examples on the R-VQA dataset. For each image-question-answer pair, the dataset provides its aligned relation fact.}
	\label{rvqa_example}
\end{figure}

\textbf{Human Evaluation~} To ensure the quality of matched facts, we employ crowdsourcing workers to label whether the generated facts are closely related to the given QA pair.
For each generated dataset with a certain threshold score, we randomly sample 1,000 examples for human labeling and ask three workers to label them.
The final accuracy for each dataset is the average accuracy obtained by the three workers.
Additionally, the workers are encouraged to label every question-answer-fact tuple in more than three seconds.
As we can see in Table \ref{tab:rvqa-data}, with the increase of relevance score threshold, the R-VQA dataset has higher matched accuracy,
together with a smaller data size. Figure \ref{rvqa_example} shows two examples in the R-VQA dataset with a score threshold value of 0.30.

\textbf{Data Analysis~} To balance the quality of matched facts and quantity of data sample, we compromise by choosing a matched score threshold value of 0.30, leading to a dataset of 198,889 samples with an average matched accuracy of 90.8\% for all question-answer-fact tuples. 
There are 5,833, 2,608, and 6,914 unique subjects, relations, and objects, respectively, covering a wide range of semantic topics.
In Table \ref{tab:rvqa-top}, we can see the distribution of the most frequent subjects, relations, objects, and facts on the generated R-VQA dataset.

% \vspace{-1mm}

\subsection{Relation Fact Detector}  \label{subsection:42} 
% tab3:rvqa-top
\begin{table}[t]
	\centering 
	%\scriptsize
	\footnotesize
	%\small
	\begin{tabular}{*8{l}} 
		\toprule	
		\multicolumn{2}{c}{Top subjects}  & \multicolumn{2}{c}{Top relations} & \multicolumn{2}{c}{Top objects}& \multicolumn{2}{c}{Top Facts}  \\
		\midrule
		\textit{man} \hspace{-2mm}   &   7.80 \%   
		&  \textit{is} \hspace{-2mm}   &   42.68 \%  
		& \textit{white} \hspace{-2mm}   &   6.69 \%  
		&   \textit{sky, is, blue} \hspace{-2mm}   &   2.22 \%  \\
		
		\textit{woman}\hspace{-2mm}    &   2.86 \%   
		& \textit{on} \hspace{-2mm}   &   21.74 \%  
		& \textit{blue} \hspace{-2mm}   &   4.08 \%  
		& \textit{grass, is, green} \hspace{-2mm}   &   1.75 \% \\
		
		\textit{sky} \hspace{-2mm}   &   2.81 \% 
		& \textit{in} \hspace{-2mm}   &   8.23 \%  
		& \textit{green} \hspace{-2mm}   &   4.02 \%  
		& \textit{cloud, in, sk} \hspace{-2mm}   &   0.76 \% \\
		
		\textit{there} \hspace{-2mm}   &   2.35 \%  
		& \textit{wearing} \hspace{-2mm}   &   3.18 \%  
		& \textit{black} \hspace{-2mm}   &   2.79 \%  
		&  \textit{plate, is, white} \hspace{-2mm}   &   0.72 \% \\
		
		\textit{grass} \hspace{-2mm}   &   2.24 \%  
		& \textit{holding} \hspace{-2mm}   &   2.71 \%  
		& \textit{red} \hspace{-2mm}   &   2.43 \%  
		& \textit{train, on, track} \hspace{-2mm}   &   0.59 \% \\
		
		\textit{cat} \hspace{-2mm}   &   2.12 \%  
		& \textit{near} \hspace{-2mm}   &   1.64 \%  
		& \textit{brown} \hspace{-2mm}   &   2.27 \% 
		& \textit{snow, is, white} \hspace{-2mm}   &   0.47 \% \\
		
		\textit{dog} \hspace{-2mm}   &   1.85 \%  
		& \textit{behind} \hspace{-2mm}   &   1.60 \%  
		& \textit{plate} \hspace{-2mm}   &   2.11 \%  
		& \textit{tree, is, green} \hspace{-2mm}   &   0.39 \% \\
		
		\textit{train}\hspace{-2mm}    &   1.50 \%  
		& \textit{above} \hspace{-2mm}   &   1.56 \%  
		& \textit{table} \hspace{-2mm}   &   2.06 \%  
		& \textit{toilet, is, white} \hspace{-2mm}   &   0.37 \% \\
		
		\textit{tree} \hspace{-2mm}   &   1.45 \%  
		& \textit{sitting on} \hspace{-2mm}   &   1.47 \%  
		&  \textit{wall} \hspace{-2mm}   &   1.61 \%  
		&  \textit{man, wearing, shirt} \hspace{-2mm}   &   0.35 \% \\
		
		\textit{plate} \hspace{-2mm}   &   1.41 \%  
		& \textit{has} \hspace{-2mm}   &   1.12 \%  
		& \textit{water} \hspace{-2mm}   &   1.58 \%  
		&  \textit{snow, on, ground} \hspace{-2mm}   &   0.34 \% \\
		\bottomrule		
	\end{tabular}
	\caption{Top relation facts in the R-VQA dataset.}
	\label{tab:rvqa-top}
\end{table}
% \vspace{-1mm}

The Relation-VQA Dataset provides 198,889 image-question-answer-fact with a matching score of 0.30. That is to say, for each image in the dataset, a question and a correct answer corresponding to the image content are provided, as well as a relation fact well supporting the question-answer data. As stated before, a relation fact describes semantic knowledge information, which benefits a VQA model a lot with better image understanding. For these reasons, we develop a relation fact detector to obtain a relation fact related to both the question and image semantic content. The fact detector will be further expanded in our relation fact-based VQA model, as illustrated in Section \ref{section:vqa}.

\textbf{Detector Modeling~} Given the input image and question, we formulate the fact prediction
as a multi-task classification following \cite{li2017vip,liang2017deep}. 
For the image embedding layer, we feed the resized image to a pre-trained ResNet-152\cite{he2016deep},
and take the output of the last convolution layer as a spatial representation of the input image content.
Then we add a spatial average pooling layer to extract a dense image representation $v \in \mathcal{R}^{2048} $  as
\begin{align}
	v= \mathrm{Meanpooling} (\mathrm{CNN}(I)) .
\end{align}
The Gated Recurrent Unit (GRU) network is adopted to encode the input question semantic feature as $q \in \mathcal{R}^{2400} $
\begin{align}
	q = \mathrm{GRU} (Q) .
\end{align}

To encode the image and question in a shared semantic space, the feature representations $v$ and $q$
are fed into a linear transformation layer followed by a non-linear activate function, respectively, as the following equations, 
\begin{align}
	f_v = \mathrm{tanh} (W_vv+b_v), ~~ f_q = \mathrm{tanh} (W_qq+b_q), 
\end{align}
where $W_v, W_b, b_v, b_q$ are the learnable parameters for linear transformation,
and $\mathrm{tanh}$ is a hyperbolic tangent function.
	
A joint semantic feature embedding is learned by combing the image and question embeddings in the common space,
\begin{align}
	h =\mathrm{tanh}(W_{vh}f_v+W_{qh}f_q+b_h) . 
\end{align}
where element-wise addition is employed for the fusion strategy of two modalities. After fusing the image and question representations, a group of linear classifiers are learned for predicting the \textit{subject}, \textit{relation} and \textit{object} in a relation fact,  
\begin{align}
	p_{sub} &= \mathrm{softmax}(W_{hs} h+b_s), \label{eq:4_1} \\
	p_{rel} &= \mathrm{softmax}(W_{hr} h+b_r), \label{eq:4_2} \\
	p_{obj} &= \mathrm{softmax}(W_{ho} h+b_o), \label{eq:4_3} 
\end{align}
where $ p_{sub}, p_{rel}, p_{obj}$ denote the classification probabilities for subject, relation and  object over pre-specific candidates, respectively. Our loss function combines the group classifiers as
\begin{align}
L_{t} = \lambda_s L(s,\hat{s}) + \lambda_r L(r,\hat{r}) +\lambda_o L(o,\hat{o}) + \lambda_w \|W\| _2,
\end{align}
where $ s, r, o $ are target subjects, relations, and objects, and $ \hat{s}, \hat{r},\hat{o} $ are the predicted results. $\lambda_s=1.0, \lambda_r=0.8, \lambda_o=1.2$ are hyper-parameters obtained though grid search on the development set.
$L$ denotes the cross entropy criterion function used for multi-class classification. An L2 regularization term is added to prevent overfitting, and the regularization weight $\lambda_w$ is set to $1\times10^{-7}$ in our experiment.

\begin{figure}[t]
	\centering
	\includegraphics[width=0.95\linewidth]{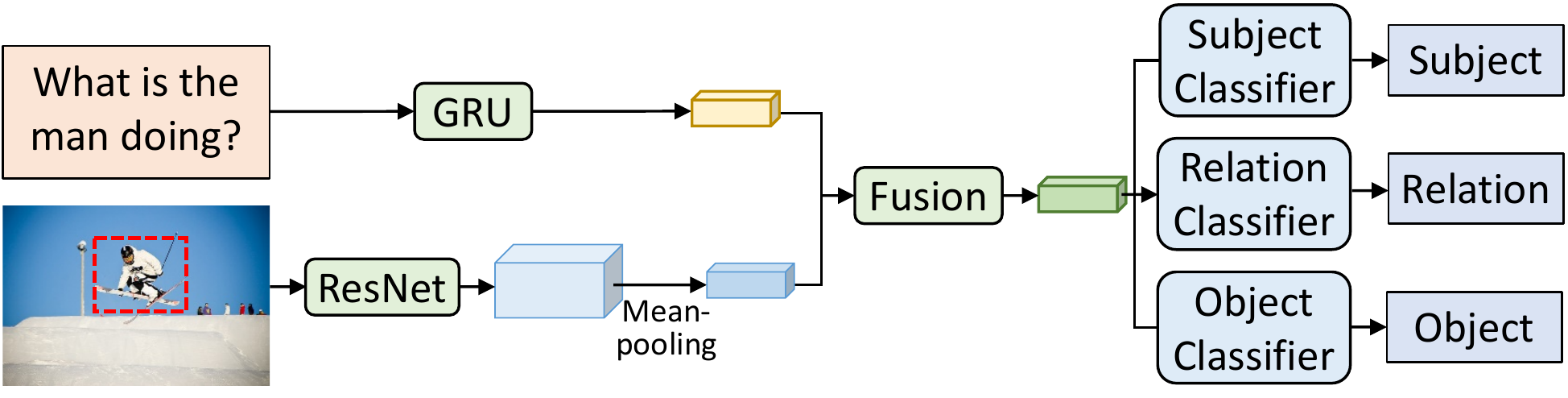}
	\caption{Relation Fact Detector.}
	\label{detector}
\end{figure}
% \vspace{-1mm}

\textbf{Experiments~} Given an input image and question, the goal of the proposed relation detector is to generate a set of relation facts
\textit{subject,relation,object} related to semantic contents of both image and question. The possibility of a predicted fact is
the sum of probabilities of the subject, relation, and object in Eqs \ref{eq:4_1}-\ref{eq:4_3}.
We conduct experiments on the training and development sets for learning, and the testing set for evaluation. 

Before carrying out the experiments, some essential operations of data preprocessing are performed. It is observed that there exist some similar and synonymous elements in facts on R-VQA, which may confuse the training of fact detection. For example, ``\textit{on}'' vs. ``\textit{on the top of}'' vs. ``\textit{is on}'', ``\textit{tree}'' vs. ``\textit{trees}'', etc. Therefore, we merge these ambiguous elements to their simplest forms based on alias concept dictionaries labeled by \cite{krishna2017visual},  e.g., ``\textit{on the top of}'' and ``\textit{is on}'' are simplified to ``\textit{on}''. The merging results are shown in Table \ref{tab:detector-data}. We take the most frequent subjects, relations, and objects from all unique candidates in training data, which leads to 2,000 subjects, 256 relations and 2,000 objects, respectively, with more details shown in Table \ref{tab:detector-data}.

The evaluation metrics we report are \textbf{recall@1}, \textbf{recall@5}, and \textbf{recall@10}, similar to \cite{lu2016visual}. \textbf{recall@k} is defined as the fraction of numbers the correct relation fact is predicted in the top \textbf{k} ranked predicted facts. The RMSProp learning method is adopted to train the detector, with an initial learning rate of $3\times10^{-4}$, a momentum of 0.98 and a weight decay of 0.01. The batch size is set to 100,
and dropout strategy is applied before every linear transformation layer. 

% tab4:detector-data
\begin{table}[t]
	\centering 
	%\scriptsize
	\footnotesize
	%\small
	\begin{tabular}{*7{c}} 
		\toprule	
		\multirow{2}{*}{}	& \multicolumn{2}{c}{Top subjects (2k)} 	& \multicolumn{2}{c}{Top relations (256)} 	& \multicolumn{2}{c}{Top objects (2k)}  \\
		\cmidrule(lr){2-3} 	\cmidrule(lr){4-5} 	  \cmidrule(lr){6-7} 	
		Operation    	&  No.   &   Perc. 		&  No.   &   Perc. 		&  No.   &   Perc.   \\
		\midrule
		Before merging    &  114,797   &   96.20   & 115,159   &   96.50		&   113,398   &   95.10  \\
		After merging  	& 115,581   &   96.86 	& 116,008   &   97.21 	& 113,962  &   95.50 \\
		\bottomrule		
	\end{tabular}
	\caption{Merging results of the R-VQA dataset for relation detector. After merging similar elements, the top element candidate in relation facts can cover more training data.}
	\label{tab:detector-data}
\end{table}
% \vspace{-1mm}

% tab5:detector-result
\begin{table}[t]
	\centering 
	\small
	\begin{tabular}{*7{c}} 
		\toprule	
		\multirow{2}{*}{}	
		& \multicolumn{3}{c}{Element (Accuracy)} 	
		& \multicolumn{3}{c}{Fact (Recall)}  \\
		\cmidrule(lr){2-4} 	\cmidrule(lr){5-7} 	
		Models  	&  Sub.   &   Rel.	&  Obj.&  R@1 & R@5	&  R@10  \\
		\midrule
		V only 		& 3.25	& 39.19	& 2.11   	& 0.14 & 0.43 & 0.72  \\
		Q only 		& 56.66	& 77.34	& 40.76 	& 23.14	& 37.82	& 43.16  \\
		\midrule
		ours - no merge 		& 65.98	& 74.79	& 43.61 	& 25.23	& 44.25	& 51.26  \\
		\textbf{ours - final}	& \textbf{66.47} & \textbf{78.80} & \textbf{45.13} 	
		& \textbf{27.39} & \textbf{46.72} & \textbf{54.10}  \\
		\bottomrule		
	\end{tabular}
	\caption{Results for the relation detector.}
	\label{tab:detector-result}
\end{table}
%\vspace{-1mm}

\textbf{Results~} Table \ref{tab:detector-result} shows the experiment results on the R-VQA test set.
The first part of Table \ref{tab:detector-result} reports two baseline models,
which fully supports that both image and question semantic information is beneficial to relation fact prediction. 
On the one hand, the model without question content (denoted as \textbf{V only}) shows a sharp drop in the accuracy of predicted facts.
This phenomenon is intuitive since semantic facts and questions both come from textual modality, while images come from visual modality.
In order to improve the semantic space of relation facts, we formulate fact prediction as a multi-objective classification problem,
and candidate facts are combinations of three elements, namely a subject, a relation, and an object.
Therefore, it is important to provide the question semantic information to reduce the space of candidate facts. 
On the other hand, the model without image content (denoted as \textbf{Q only}) suffers from limited prediction performance,
indicating images also contain some useful semantic knowledge.

The second part of Table \ref{tab:detector-result} illustrates that the model based on the merged R-VQA data (denoted as \textbf{Ours - final})
works much better than the model  based on initial R-VQA data (denoted as \textbf{Ours - no merge}). 
Although existing methods have made good progress in visual detection achieving \textbf{Rec@100} accuracy of 10-15\% on Visual Genome for visual facts, these approaches are not suitable to predict question-related visual facts.
In contrast with these works, our model incorporates the question feature for fact prediction,
and achieves a much higher accuracy with a smaller candidate number of \textbf{k}, as well as a much simpler framework.
In future work, it will be still meaningful to design a fine-grained model to obtain better prediction performance.

\begin{figure*}[t]
	\centering
	\includegraphics[width=0.9\linewidth]{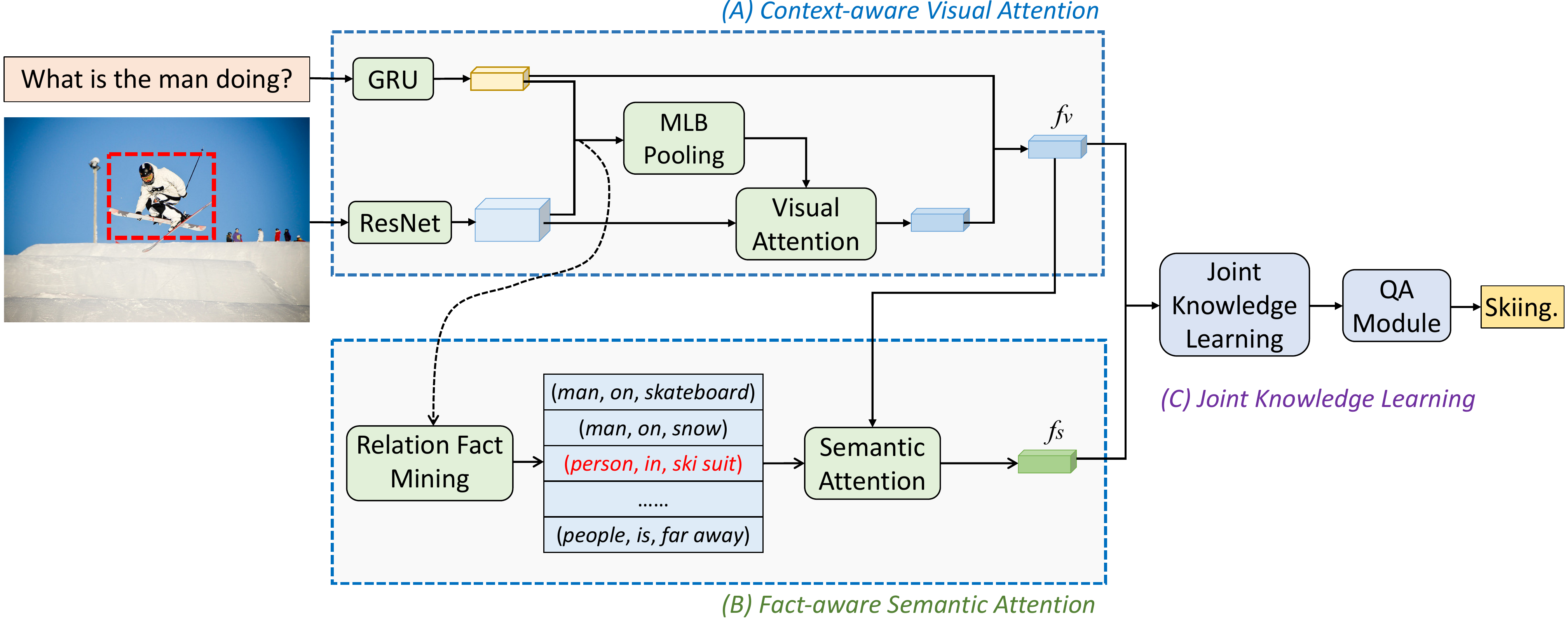}
	\caption{Our proposed multi-step attention network for VQA.}
	\label{vqamodel}
\end{figure*}
% \vspace{-1mm}

%%%%%%%%%%%%%%%%%%%%%%%%%%%%%%%%%%%%%%%%%%%%%%%%%%%%%%%%%%%%%%%%%%%%%%%

\section{Visual Question Answering with Facts}\label{section:vqa}

The overall framework of our proposed multi-step attention network for VQA is demonstrated in Figure \ref{vqamodel}, which takes a semantic question and an image as inputs, and learns visual and semantic knowledge sequentially to infer the correct answer. Our proposed network consists of three major components: (A) Context-aware Visual Attention, (B) Fact-aware Semantic Attention, and (C) Joint Knowledge Embedding Learning. Context-aware visual attention is designed to select image regions associated with the input question and to obtain visual semantic representation of these regions. Fact-aware semantic attention aims to weigh detected relevant relation facts by the learned visual semantic representation, and to learn semantic knowledge.
Finally, a joint knowledge embedding learning model is able to jointly encode visual and semantic knowledge and infer the most possible answer.

\subsection{Context-aware Visual Attention}\label{subsection:51} 

Similar to many previous VQA approaches \cite{fukui2016multimodal,xu2016ask,yu2017multi}, we adopt
a question-aware visual attention mechanism to choose related image regions. 

\textbf{Image Encoding~} We apply a ResNet-152 network \cite{he2016deep} to extract image feature embedding for an input image. The $2048\times14\times14$ feature map from the last convolution layer is taken as the image visual feature $v$, which corresponds to $14\times14$ image regions with $2048$ feature channels.

\textbf{Question Encoding~} A gate recurrent unit (GRU) \cite{cho2014learning} is used to encode the question embedding, which is widely adopted in NLP and multimodal tasks \cite{long2017cognition,li2017context,yu2017multi}. To be specific, given a question with $T$ words $Q = [q_1,...,q_t,...,q_T]$, where $q_t$ is the one hot vector of the question word at position $t$, we first embed them into a dense representation via a linear transformation $x_t = W_eq_t$. At each time $t$, we feed the word embedding $x_t$ into the GRU layer sequentially, and the GRU recursively updates the current hidden state $h_t = \mathrm{GRU}(h_{t-1},x_t)$ with the input $x_t$ and previous hidden state $h_{t-1}$. Finally, we take the last hidden state $h_T$ as the question representation.

\textbf{Visual Attention~} A visual attention mechanism is adopted to highlight image regions related to question semantic information, and to learn more effective multimodal features between textual and visual semantic information. First, we apply the multimodal low-rank bilinear pooling (MLB) method \cite{kim2016hadamard} to merge two modalities of the question and image as
\begin{align}
	c = \mathrm{MLB}(q, v) .
\end{align}
where context vector $c$ contains both question and image semantic content.
We map the  context vector to attention weights via a linear transformation layer followed by a softmax layer, 
\begin{align}
m = \mathrm{softmax}(W_c c +b_c) ,
\end{align}
where weights $m$ has a size of $14\times14$, and the value of each dimension represents the semantic relevance between corresponding image region and the input question. The context-aware visual feature is calculated as weighted sum of representations over all image regions, which is given by:
\begin{align}
	\tilde{v} =\sum_{i=1}^{14\times14} m(i) v(i) .
\end{align}
We further combine the context-aware visual feature with the question feature to obtain the final visual representation as
\begin{align}
	f_v = \tilde{v} \circ \mathrm{tanh}(W_q q +b_q), 
\end{align}
where $\circ$ denotes element-wise multiplication.

%%% 5-2
\subsection{Fact-aware Semantic Attention}\label{subsection:52}

Visual attention enables the mining of visual context-aware knowledge, such as object and spatial information, which is beneficial to questions mainly focusing on object detection. However, models only with visual attention may suffer from limited performance when more relation reasoning is required.
Therefore, we incorporate a list of relation facts as semantic clues and 
propose a semantic attention model to weigh different relation facts for better answer prediction. 
Some existing studies mine semantic concepts or attributes as semantic knowledge to assist VQA models. Our proposed model differs from these works in two ways. On one hand, existing methods only mine concepts or attributes, while our model extracts relation facts containing concepts and attributes, obviously increasing the semantic capacity of the semantic knowledge used. On the other hand, concepts or attributes in previous works may be irrelevant to VQA, because they are extracted only considering image content and based on data or pre-trained CNN models from other tasks like caption and object recognition \cite{yu2017multi}. 
In contrast, we build up the Relation-VQA dataset to train the relation fact detector directly focusing on both the input image and question.

\textbf{Fact Detection~} First, we incorporate the fact detector introduced previously in Section \ref{section:4} into our VQA model. Given the input image and question, the fact detector is used to generate the most possible $K$ relation facts as a candidate set $T = [t_1; t_2; ...; t_K]$. For a fact $t_i = (s_i, r_i, o_i)$, we embed each element of the fact into a common semantic space $\mathcal{R}^n$, and concatenate these three embeddings as the fact embedding as follows:
\begin{align}
	f_{t_i} = [W_{sh}s_i, W_{rh}r_i, W_{oh}o_i] \in \mathcal{R}^{3n}. 
\end{align}
Then we can obtain the representation of $K$ fact candidate, denoted as $f_T = [f_{t_1}; f_{t_2}; ...; f_{t_K}] \in \mathcal{R}^{K\times3n}$.

\textbf{Semantic Attention~} Second, we develop a semantic attention to find out important facts considering the input image and question. Concretely, we use the context-aware visual representation as a query to select significant facts in a candidate set. Similar to context-aware visual attention, given the context-aware visual embedding $f_v$ and fact embedding $f_T$, we first obtain joint context representation $c_t$ and
then calculate attention weight vector $m_t$ as follows:	
\begin{align}
	c_t &= \mathrm{MLB}(f_v, f_T) , \label{eq:sem1} \\
	m_t &= \mathrm{softmax}(W_{c_t} c_t +b_{c_t}). \label{eq:sem2}
\end{align}
The final attended fact representation over candidate facts is calculated as
\begin{align}
	f_s	&= \sum_{i=1}^K m_t(i) f_T(i)  \label{eq:sem3},
\end{align}
which serves as semantic knowledge information for answering visual questions.

%%% 5-3
\subsection{Joint Knowledge Embedding Learning} \label{subsection:53}
Our proposed multi-step attention model consists of two attention components. One is visual attention which aims to select related image regions and output context-ware visual knowledge representation $f_v$. Another is semantic attention which focuses on choosing related relation facts and output fact-ware semantic knowledge representation $f_s$. We merge these two representations via element-wise addition with linear transformation and a non-linear
activation function to jointly learn visual and semantic knowledge, 
\begin{align}
	h = \tanh(W_{vh}f_v+b_{v}) + \tanh(W_{sh}f_s+b_{s}) \label{eq:joint1}.
\end{align}
As we formulate VQA as a multi-class classification task, a linear classifier is trained to infer the final answer, 
\begin{align}
	p_{ans} = \mathrm{softmax}(W_ah+b_{a}) \label{eq:joint2}.
\end{align}

%%%%%%%%%%%%%%%%%%%%%%%%%%%%%%%%%%%%%%%%%%%%%%%%%%%%%%%%%%%%%%
\section{Experiments}
\subsection{Datasets and Evaluation Metrics}

We evaluate our proposed model on two popular benchmark datasets, th VQA dataset \cite{antol2015vqa} and the COCO-QA dataset \cite{ren2015exploring}, due to large data sizes and various question types. 

%\textbf{VQA~} 
The VQA dataset is annotated by Amazon Mechanical Turk (AMT), and  contains 248,349 training instances, 121,512 validation instances and 244,302 testing instances based on a number of 123,287 unique natural images. The dataset is made up of three question categories including \textit{yes/no}, \textit{number} and \textit{other}. For each question, ten answers are provided by different annotators. We take the top 2,000 most frequent answers following previous work \cite{kim2016hadamard} as candidate answer outputs, which cover 90.45\% of answers in training and validation sets. For testing, we train our model on the train+val set and report the testing result on the test-dev set from a VQA evaluation server maintained by \cite{antol2015vqa}. There are two different tasks, an open-ended task and a multi-choice task. For the open-ended task, we select the most possible answer from our candidate answer set, while for the multi-choice task, we choose the answer with the highest activation score among the given choices.

%\textbf{COCO-QA~} 
The COCO-QA dataset is another benchmark dataset, including 78,736 training questions and 38,948 testing questions. There are four question types, \textit{object}, \textit{number}, \textit{color}, and \textit{location}, which cover 70\%, 7\%, 17\% and 6\% of total question-answer pairs, respectively. All of the answers in the dataset are single words. As the COCO-QA dataset is smaller, we select all the unique answers as possible answers, which leads to a candidate set with a size of 430.

\textbf{Evaluation Metric~} 
For the VQA dataset, we report the results following the evaluation metric provided by the authors of the dataset, where a predicted answer is considered correct only if more than three annotators vote for that answer, that is to say, 
\begin{align}
	Acc(ans) = \min (1 , \frac{\text{\#humans vote for \textit{ans}} }{3}  ).
\end{align}
For the COCO-QA dataset, a predicted answer is regarded as correct if it is the same as the labeled answer in the dataset.

\subsection{Implementation Details}

For encoding question, the embedding size for each word is set to 620. For encoding facts in the VQA model, the top ten facts are generated and the size of element embedding size $m$ is set as 900.  All other visual and textual representations are vectors of size 2400.

We implement our model with the Torch computing framework, one of the most popular recent deep learning libraries. In our experiments, we utilize the RMSProp method for the training process with mini-batches of 200, an initial learning rate of $3\times 10^{-4}$, a momentum of 0.99, and a weight-decay of $10^{-8}$. The validation is performed every 10,000 iterations and early stopping is applied if the validation accuracy does not improve at the last five validations. We use a drop strategy with a probability of 0.5 at every linear transformation layer to reduce overfitting.

\subsection{Comparison with State-of-the-art}

\begin{table}[t]
	\centering 
	%\scriptsize
	\footnotesize
	%\small 
	\begin{tabular}{{l}*{8}{@{\hspace{2.8mm}}c}}
		\toprule
		\multirow{2}{*}{} &  \multicolumn{4}{c}{Open-Ended} &	\multicolumn{4}{c}{Multi-Choice}	\\
		\cmidrule(lr){2-5} 	\cmidrule(lr){6-9} 
		\textbf{Method}	& All & Y/N &  Num.  & Other  & All & Y/N & Num. & Other \\	
		
		\midrule
		LSTM Q+I \cite{antol2015vqa}  	 
		& 53.74 & 78.94 & 35.24 & 36.42 	& 57.17	& 78.85 & 35.80 & 43.41  \\
		DPPnet \cite{noh2016image}	
		& 57.22 & 80.71 & 37.24 & 41.69 	& 62.48 & 80.79 & 38.94 & 52.16   \\
		FDA \cite{ilievski2016focused}  
		& 59.24 & 81.14 & 36.16 & 45.77 	& 64.01 & 81.50 & 39.00 & 54.72  \\
		DMN+ \cite{xiong2016dynamic} 
		& 60.30 & 80.50 & 36.80 & 48.30 	& -	& - & - & -  \\
		
		\midrule
		SMem \cite{xu2016ask}
		& 57.99 &  80.87 & 37.32 & 43.12 	& -	& - & - & -   \\
		SAN \cite{yang2016stacked}		  	
		& 58.70 & 79.30 & 36.60 & 46.10  	& -	& - & - & - \\
		QRU \cite{li2016visual}				
		& 60.72 & 82.29 & 37.02 & 47.67	& 65.43 & 82.24 & 38.69 & 57.12  \\
		MRN \cite{kim2016multimodal}
		& 61.68 & 82.28 & 38.82 & 49.25 	& 66.15	& 82.30 & \textbf{40.45} & 58.16  \\
		% HieCoAtt \cite{lu2016hierarchical} 
		% & 61.80 & 79.70 & 38.70 & 51.70 	& 65.80	& 79.70 & 40.00 & 59.80    \\
		MCB \cite{fukui2016multimodal}
		& 64.20 & 82.20 & 37.70 & 54.80 	& 68.60	& - & - & -   \\
		MLB \cite{kim2016hadamard}
		& 64.53 & 83.41 & 37.82 & 54.43     & -	& - & - & -   \\
			
		\midrule
		V2L \cite{wu2016value}	 	  		
		& 57.46 & 78.90  & 36.11 & 40.07 	& -	& - & - & -  \\ 
		AMA \cite{wu2016value}
		& 59.17 &  81.01 & 38.42 & 45.23 	& -	& - & - & -   \\	
		MLAN \cite{yu2017multi}
		& 64.60 &  83.80 & \textbf{40.20} & 53.70 	& 64.80	& - & - & -   \\		
		\midrule
		\textbf{RelAtt} (ours) 	
		& \textbf{65.69}	& \textbf{83.55} 	& 36.92 & \textbf{56.94}  
		& \textbf{69.60}	& \textbf{83.58} & 38.56 & \textbf{64.65}   \\	
		\bottomrule	
	\end{tabular}
	\caption{Evaluation results for our proposed model and compared methods on the VQA dataset.}
	\label{tab:vqa}
\end{table}

\begin{table}[t]
	\centering 
	% \small
	\begin{tabular}{{l}*7{c}} 
		\toprule
		\textbf{Method}	 & All  & Obj. & Num. & Color & Loc.  \\
		\midrule
		2VIS+BLSTM \cite{ren2015exploring}	& 55.09 & 58.17 & 44.79 & 49.53 & 47.34  \\	
		IMG-CNN \cite{ma2016learning}		& 58.40 & - & - & - & -	\\
		DDPnet \cite{noh2016image}	  		& 61.16	& - & -  & - & - \\
		SAN \cite{yang2016stacked}	  		& 61.60 & 65.40  & 48.60 & 57.90 & 54.00  \\
		\midrule		 	 		 
		AMA \cite{wu2016ask}	 	  		& 61.38 & 63.92  & 51.83 & 57.29 & 54.84 \\ 
		QRU \cite{li2016visual}	 	  		& 62.50 & 65.06  & 46.90 & 60.50 & 56.99  \\
		\midrule
		\textbf{RelAtt} (ours)    
		& \textbf{65.15} &\textbf{67.50} &\textbf{48.81} &\textbf{62.64} &\textbf{58.37}   \\		 
		\bottomrule
	
	\end{tabular}
	\caption{Evaluation results for our proposed model and compared methods on the COCO QA dataset.}
	\label{tab:coco}
\end{table}

Table \ref{tab:vqa} demonstrates our proposed model for both open-ended and multi-choice tasks with state-of-the-arts on the VQA test set. 
Note that all listed approaches apply only one type of visual feature and the report results of a single model.

The first part in the table shows models using simple multimodal joint learning without an attention mechanism. Models in the second part are based on visual attention, while models in the third part apply semantic attention to learn semantic knowledge like concepts and attributes. It's shown that our proposed multi-step semantic attention network (denoted as \textbf{RelAtt}) improves the state-of-the-art \textbf{MLAN} \cite{yu2017multi} model from 64.60\% to 65.69\% on the open-ended task, and from 64.80\% to 69.60\% on the multi-choice task. To be specific, our model obtains the improvement of 2.51\% in the question types \textit{Other}. As the state-of-the-art model, apart form visual attention, \textbf{MLAN} uses semantic attention to mine important concepts based on image content.
In contrast, our model \textbf{RelAtt} introduces relation facts instead of concepts as semantic knowledge, which obviously increase semantic capacity.
Moreover, we train a relation detector to learn facts based on both visual and textual content, instead of only using the image \cite{yu2017multi}.	
As our proposed R-VQA dataset extended from Visual Genome dataset shares similar image semantic space with current 
datasetes like VQA and COCO-QA, semantic knowledge learned from the fact detector can be easily transferred to the VQA task.
These are the main reasons that \textbf{RelAtt} beats \textbf{MLAN} significantly.

Table  \ref{tab:coco} compares our approach with state-of-the-arts on the COCO-QA dataset. 
Different from the VQA dataset, COCO-QA doesn't contain the multi-choice task, and fewer results are reported on it.
Our model improves the state-of-the art \textbf{QRU} \cite{li2016visual} from 62.50\% to 65.15\% with a growth of 2.65\%. In particular, our model significantly outperforms the state-of-the-art semantic attention model \textbf{AMA} \cite{wu2016ask} by 3.77\%, indicating the benefits of modeling semantic relation facts
and learning semantic knowledge from R-VQA dataset.

\subsection{Ablation Study}

In this section, we conduct five ablation experiments to study the role of individual components designed in our model. Table \ref{tab:val} reports the ablation results of compared baseline models, which are trained on the training set, and evaluated on the validation set. Specifically, the ablation experiments are as follows:
\begin{itemize}
	\item \textit{Q+I}, where we only take the image and question to infer the answer, and image-question joint representation is learned by a simple fusion method of element wise addition.
	\item \textit{Q+R}, where only the question and relation facts generated by the detector are considered to predict the answer.
	\item \textit{Q+I+Att}, where we apply visual attention to learn the joint representation of the image and question.
	\item \textit{RelAtt-Average}, where the semantic attention mechanism denoted in Eqs \ref{eq:sem1} - \ref{eq:sem3} is removed from our best model \textbf{RelAtt}. Instead, the fact representation is calculated by averaging different fact embeddings.
	% \item \textit{RelAtt-V query}, where 
	\item \textit{RelAtt-MUL}, where element-wise addition is replaced by multiplication in Eq \ref{eq:joint1} to learn the joint knowledge embedding.
\end{itemize}

The results of first three ablated models indicate that visual attention provides limited visual information for question answering and relation facts can play an important role as they contain semantic information. A drop of 0.79\% in accuracy for \textit{RelAtt-Average} illustrates that semantic attention is essential to encode relation facts. Moreover, it is shown that the fusion method of element-wise addition might work better than multiplication when encoding joint visual-textual knowledge representation.

\begin{figure}[t]
	\centering
	\includegraphics[width=0.95\linewidth]{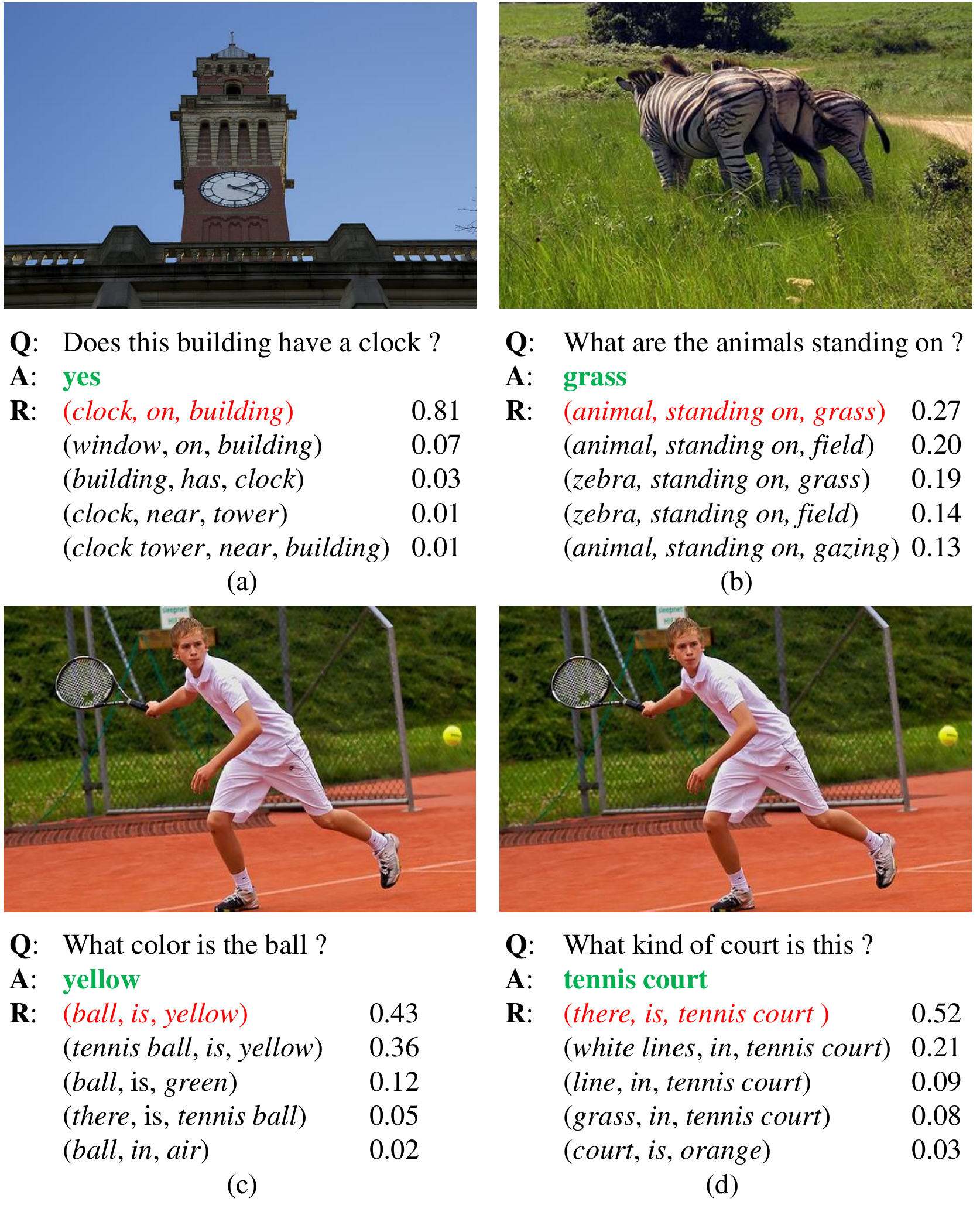}
	\caption{Testing samples on the VQA test set.}
	\label{fig5-vis}
\end{figure}

\begin{table}[t]
	\centering 	
	\begin{tabular}{{l}*{5}{c}}
		\toprule
		\textbf{Method}  & \textbf{Accuracy}  \\
		\midrule
		Q+I	 			& 53.22  \\	
		Q+R	 			& 51.34  \\ 
		Q+I+Att	 		& 57.40  \\	
		\midrule
		RelAtt-Average	& 57.84  \\
		% RelAtt-V query	& 57.64  \\ 
		RelAtt-MUL		& 58.12  \\
		\midrule
		\textbf{RelAtt} (final)	& \textbf{58.63}  \\	
		\bottomrule
	\end{tabular}
	\caption{Ablation study on the VQA dataset.}
	\label{tab:val}
\end{table}
% \vspace{-1mm}

\subsection{Case Study}

To illustrate the capability of our model in learning relation facts as semantic knowledge, we show some examples on the VQA testing set with the image, question and predicted answer. We also list relation facts generated by the fact detector and their attention 
weights in the semantic attention component. For saving space, only five in ten relation facts are shown in Figure \ref{fig5-vis}. In Figures \ref{fig5-vis} (a) and (b), the fact detector mines semantic fact candidates related to both the image and the question, and semantic attention highlights the most possible facts for question answering. In Figures \ref{fig5-vis} (c) and (d), although given the same image, the fact detector can depend on the different questions to generate corresponding semantic facts.

\section{Conclusion}
In this paper, we aim to learn visual relation facts from images and questions for semantic reasoning of visual question answering.
We propose a novel framework by first learning a relation factor detector based on the built Relation-VQA (R-VQA) dataset.
Then a multi-step attention model is developed to incorporate the detected relation facts with sequential visual and semantic attentions,
enabling the effective fusion of visual and semantic knowledge for answering.
Our comprehensive experiments show our method outperforms state-of-the-art approaches and demonstrate the effectiveness
of considering visual semantic knowledge.

\begin{acks}
%This work was supported in part by National Natural Science Foundation of China under Grant No. 61532010 and No. 61702190, National Basic Research Program of China (973 Program) under Grant No. 2014CB340505, and SHMEC (16CG24),
%in part by Microsoft Corpation.
We would like to thank our anonymous reviewers for their constructive feedback and suggestions.
This work was supported in part by the National Natural Science Foundation of China under Grant No. 61532010 and No. 61702190, 
in part by the National Basic Research Program of China (973 Program) under Grant No. 2014C\\B340505, 
in part by the Shanghai Sailing Program under Grant No. 17YF1404500,  
in part by the Shanghai Chenguang Program under Grant No. 16CG24, and 
in part by  Microsoft Corporation.
\end{acks}

% \clearpage  % [PAN] New page
% \nocite{*}  % [PAN] Include all references
\bibliographystyle{ACM-Reference-Format}
\balance
\bibliography{bibliography}

\end{document}